\documentclass{article}
\usepackage{acra}
\usepackage{graphicx} 
\usepackage{color, colortbl}
\usepackage{subfigure}
\usepackage{epsfig} 
\usepackage{amsmath} 
\usepackage{amssymb}  
\usepackage[linesnumbered, lined, ruled]{algorithm2e}
\usepackage{epstopdf}
\usepackage{bm}
\usepackage{hyperref}
\usepackage{url}
\usepackage{stfloats}
\usepackage{array}
\usepackage[T1]{fontenc}
\usepackage{booktabs}
\usepackage{lastpage}
\usepackage{multirow}
\usepackage{todonotes}
\usepackage{wrapfig}

\definecolor{Gray}{gray}{0.7}

\def\fps@figure{htp}
\def\fps@table{htp}

\newcommand{\bfig}{\begin{figure}}
\newcommand{\efig}{\end{figure}}

\newcommand{\benum}{\begin{enumerate}}
\newcommand{\eenum}{\end{enumerate}}

\newcommand{\ba}{\begin{eqnarray}}
\newcommand{\ea}{\end{eqnarray}}

\newcommand{\unit}[1]{\mbox{$\rm \,#1$}}

\title{Modular Deep Q Networks for Sim-to-real Transfer\\ of Visuo-motor Policies}

\author{
  Fangyi Zhang, J\"urgen Leitner, Michael Milford, Peter Corke\\
  Australian Centre for Robotic Vision (ACRV), Queensland University of Technology (QUT)\\ 
  Brisbane, Australia\\
  \texttt{fangyi.zhang@hdr.qut.edu.au} \\
}

\begin{document}

\maketitle



\begin{abstract}

While deep learning has had significant successes in computer vision thanks to the abundance of visual data, collecting sufficiently large real-world datasets for robot learning can be costly. 
To increase the practicality of these techniques on real robots, we propose a modular deep reinforcement learning method capable of transferring models trained in simulation to a real-world robotic task. 
We introduce a bottleneck between perception and control, enabling the networks to be trained independently, but then merged and fine-tuned in an end-to-end manner to further improve hand-eye coordination. On a canonical, planar visually-guided robot reaching task a fine-tuned accuracy of 1.6 pixels is achieved, a significant improvement over naive transfer (17.5 pixels), showing the potential for more complicated and broader applications. 
Our method provides a technique for more efficient learning and transfer of visuo-motor policies for real robotic systems without relying entirely on large real-world robot datasets.
    
    %

\end{abstract}

\section{Introduction}
\label{sec:intro}

The advent of large datasets and sophisticated machine learning models, commonly referred to as deep learning, has in recent years created a trend away from hand-crafted solutions towards more data-driven ones.
Learning techniques have shown significant improvements in robustness and performance~\cite{krizhevsky2012imagenet}, particularly in the computer vision field.
Traditionally robotic reaching approaches have been based on crafted controllers that combine (heuristic) motion planners with the use of hand-crafted features to localize the target visually. Recently learning approaches to tackle this problem have been presented~\cite{zhang2015towards,levine2016learning,bateux2017visual,katyal2017leveraging}.
However a consistent issue faced by most approaches is the reliance on large amounts of data to train these models.
For example, Google researchers addressed this problem by developing an "arm farm" with 6 to 14 robots collecting data in parallel~\cite{levine2016learning}.
Generalization forms another challenge: many current systems are brittle when learned models are applied to robotic configurations that differ from those used in training.
This leads to the question: 
\textit{is there a better way to learn and transfer visuo-motor policies on robots for tasks such as reaching?}

\begin{figure}[tpb!]
\begin{center}
\includegraphics[width=1.0\columnwidth]{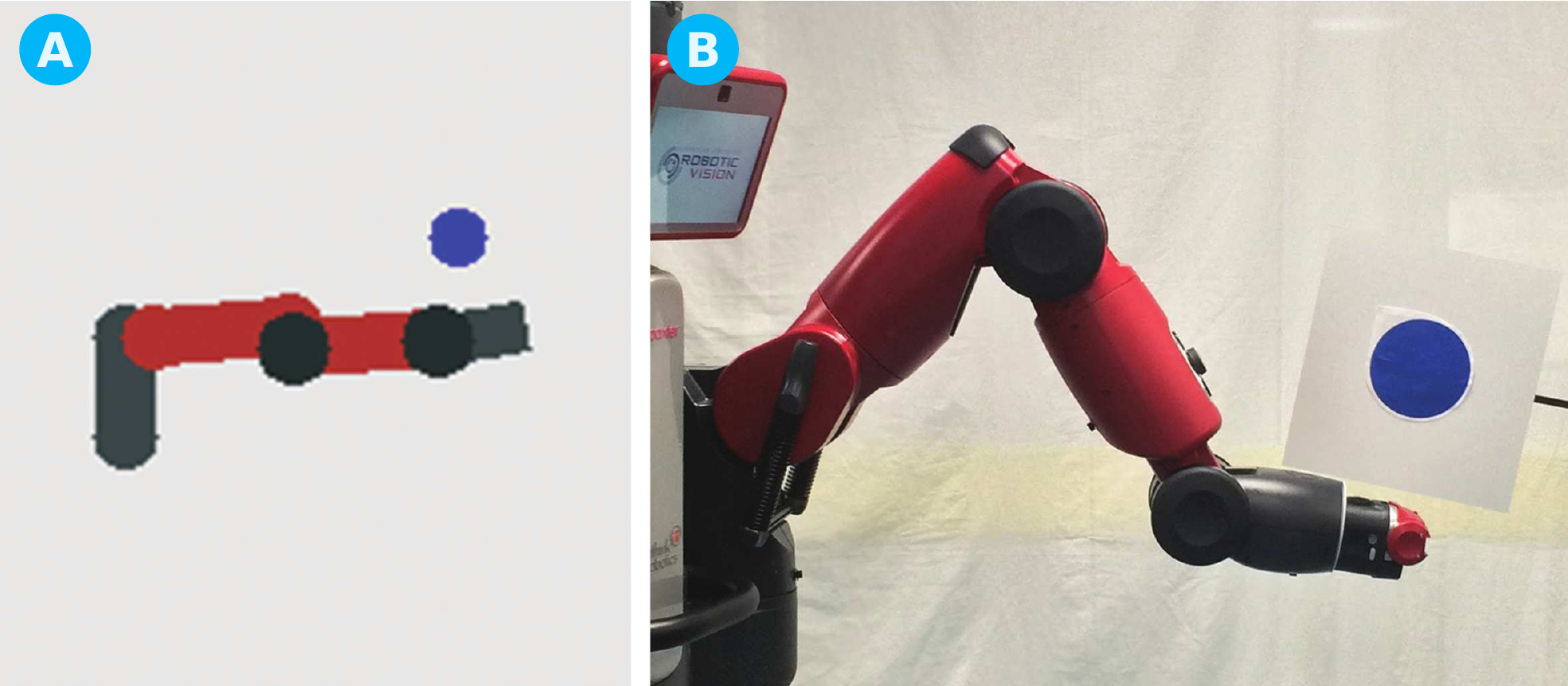}\\
\vspace{2mm}
\includegraphics[width=1.0\columnwidth]{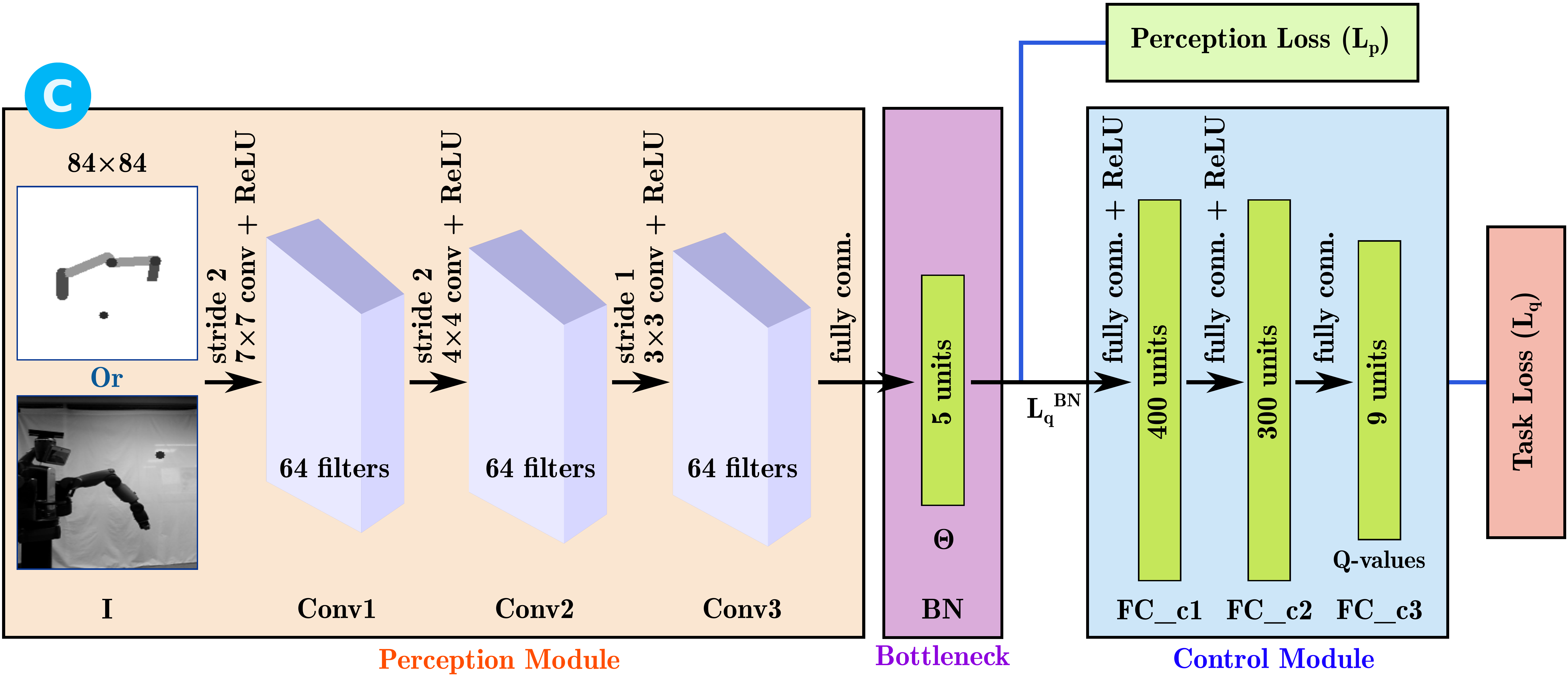}
\end{center}
\caption{%
We present a technique for efficient learning and transfer of visuo-motor policies for a planar reaching task from simulated (A) to real environments (B) using a modular deep Q network (C).%
} 
\label{fig:sim_to_real}
\end{figure}

Various approaches have been proposed to address these problems in a robot learning context:
\textbf{(i)} the use of simulators, simulated and synthetic data \cite{bateux2017visual,d2017bridging,tobin2017domain,james2017transferring};
\textbf{(ii)} methods that transfer the learned models to real-world scenarios \cite{fitzgerald2015similarity,tzeng2016adapting};
\textbf{(iii)} directly learning real-world tasks by collecting large amounts of data \cite{levine2016learning,pinto2016supersizing}.

In this paper, we present a method that connects these three, usually separately considered, approaches. Vision and kinematics data is gathered in simulations (cheap) to decrease the amount of real world collection necessary (costly). The approach is capable of transferring the learned models to real-world scenarios with a fraction of the real-world data typically required for direct real-world learning approaches.

In particular, we propose a modular deep reinforcement learning approach -- inspired by DQN for Atari game playing~\cite{mnih2015human} -- to efficiently learn and transfer visuo-motor policies from simulated to real environments, and benchmark with a visually-guided planar reaching task for a robotic arm (Figure~\ref{fig:sim_to_real}).
By introducing a modular approach, the perception skill and the controller can be transferred individually to a robotic platform, while retaining the ability to fine-tune them in an end-to-end fashion to further improve hand-eye coordination on a real robot (in this research a Baxter).

\section{Related Work}
\label{sec:related_work}

Data-driven learning approaches have become popular in computer vision and are starting to replace hand-crafted solutions also in robotic applications.
Especially robotic vision tasks -- robotic tasks based directly on real image data -- such as,  navigation~\cite{taideep},
object grasping and manipulation~\cite{levine2016learning,pinto2016supersizing,lenz2015deepmpc} have seen increased interest.
The lack of large-scale real-world datasets, which are expensive, slow to acquire and limit the general applicability of the approach, has so far limited the broader application.
Collecting the datasets required for deep learning has been sped up by using many robots operating
in parallel~\cite{levine2016learning}. With over 800,000 grasp attempts recorded, a deep network was trained to predict the success probability of a sequence of motions aiming at grasping on a 7 DoF robotic manipulator with a 2-finger gripper.  
Combined with a simple derivative-free optimization algorithm 
the grasping system achieved a success rate of 80\%. Another example of dataset collection for grasping is the approach to self-supervised grasping learning in the real world where force sensors were used to autonomously label samples~\cite{pinto2016supersizing}. After training with 50,000 real-world trials using a staged leaning method, a deep convolutional neural network (CNN) achieved a grasping success rate around 70\%.
These are impressive results but achieved at high cost in terms of dollars, space and time.


DeepMind showed that a deep reinforcement learning system is able to directly synthesize control actions for computer games from vision data~\cite{mnih2015human}. 
While this result is an important and exciting breakthrough it does not transfer directly to real robots with real cameras observing real scenes~\cite{zhang2015towards}.
In fact very modest image distortions in the simulation environment (small translations, Gaussian noise and scaling of the RGB color channels) caused the performance of the system to fall dramatically. Introducing a real camera observing the 
game screen was even worse~\cite{tow2016robustness}.

There has been increasing interest to create robust visuo-motor policies for robotic applications, especially in reaching and grasping.
Levine~et~al.\ introduced a CNN-based policy representation architecture with an added guided policy search (GPS) to learn visuo-motor policies (from joint angles and camera images to joint torques)~\cite{levine2016end}, which allow to reduce the number of real world training by providing an oracle (or expert's initial condition to start learning).
Impressive results were achieved in complex tasks, such as hanging a coat hanger, inserting a block into a toy, and tightening a bottle cap.
Recently it has been proposed to simulate depth images to learn and then transfer grasping skills to real-world robotic arms \cite{viereck2017learning}, yet no adaptation in the real-world has been performed.


Transfer learning attempts to develop methods to transfer knowledge between different tasks~\cite{taylor2009transfer,pan2010survey}. To reduce the amount of data collected in the real world (expensive), transferring skills from simulation to the real world is an attractive alternative.
Progressive neural networks are leveraged to improve transfer and avoid catastrophic forgetting when learning complex sequences of tasks~\cite{rusu2016progressive}.
Their effectiveness has been validated on reinforcement learning tasks, such as Atari and 3D maze game playing.
Modular reinforcement learning approaches have shown skill transfer capabilities in simulation~\cite{devin2016learning}.
However, methods for real-world robotic applications are still scarce and require manually designed mapping information, e.g.\ similarity-based approach to skill transfer for robots~\cite{fitzgerald2015similarity}. 
To reduce the number of real-world images required,
 a method of adapting visual representations from simulated to real environments was proposed, achieving a success rate of 79.2\% in a ``hook loop'' task, with 10 times less real-world images~\cite{tzeng2016adapting}.

\section{Methodology}
\label{sec:methodology}

Reinforcement learning~\cite{barto1998reinforcement} has been proposed for agents to learn novel behaviours.
One approach for learning from rewards is Q-learning~\cite{sutton1998reinforcement}, which aims to obtain a policy $\pi$ that maximizes the expectation of accumulated rewards by approximating an optimal Q-value function
\begin{equation}
\label{equ:q_learning}
Q^*(s,a)=\underset{\pi}{\max}\ \mathbb{E}\Big[ \sum_{i=0}^{\infty} \gamma^{i} r_{t+i} | s_t=s, a_t=a, \pi \Big],
\end{equation}
where $r_t$ is the reward at each time step $t$, when following a behaviour policy $\pi=P(a|s)$ that determines which action $a$ to take in each state $s$.
$\gamma$ is a discount factor applied to future rewards. 
A deep neural network was introduced to approximate the Q-value function, named Deep Q Network (DQN)~\cite{mnih2015human}.
The state can therefore be represented by a high-dimensional raw-pixel image, since latent state features can be extracted by the convolutional layers~\cite{krizhevsky2012imagenet}. However, learned visuo-motor policies with high-level (raw pixel) input do not transfer directly from simulated to real robots~\cite{zhang2015towards}.


\subsection{Modular Deep Q Networks}


\label{sec:modular_net}

Our preliminary studies of deep visuo-motor policies indicate that the convolutional layers focus on perception, i.e., extracting useful information from visual inputs, while the fully connected (FC) layers perform control~\cite{zhang2017cvpr}. 
To make the learning and transfer of perception and control more efficient, we propose to separate the DQN into perception and control modules connected by a bottleneck layer (Figure~\ref{fig:sim_to_real}C).
The \textbf{bottleneck} forces the network to learn a low-dimensional representation, not unlike Auto-encoders~\cite{hinton2006reducing}. The difference is that we  \textbf{explicitly equate} the bottleneck layer with the minimal \textbf{scene configuration} $\mathbf{\Theta}$ whose meaning will be further introduced in Section~\ref{sec:benchmark}.
The values in $\mathbf{\Theta}$ are normalized to the interval $[0,1]$.

With the bottleneck, the perception module learns how to estimate the scene configuration $\hat{\mathbf{\Theta}}$ from a raw-pixel image $I$; the control module learns to approximate the optimal Q-value function as defined in Eq.~\ref{equ:q_learning}, determining the most appropriate action $a^*$ given the scene configuration $\mathbf{\Theta}$, i.e., $a^* = \underset{a}{\max}\ Q(\mathbf{\Theta},a)$.

To further improve the performance of a combined network (perception + control), a weighted end-to-end fine-tuning method is proposed, since experimental results show that
a naive end-to-end fine-tuning using a straight-forward loss function does not work well for performance improvement (Section~\ref{sec:e2e_reaching}).

\subsection{Training Method}


\subsubsection{Perception}
The perception network is trained using supervised learning -- first conducted in simulation, then fine-tuned with a small number of real samples for skill transfer -- with the quadratic loss function
\begin{equation}
\label{equ:perception_cost}
L_p=\frac{1}{2m} \sum_{j=1}^{m} \left \| y(I^j) - \mathbf{\Theta}^j \right \|^2,
\end{equation}
where $y(I^j)$ is the prediction of $\mathbf{\Theta}^j$ for $I^j$; $m$ is the number of samples.

\subsubsection{Control}
The control network is trained using Q-learning, where weights are updated using the Bellman equation which is equivalent to the loss function 
\begin{equation}
\label{equ:control_cost}
L_q=\frac{1}{2m} \sum_{j=1}^{m} \left \| Q(\mathbf{\Theta}_t^j,a_t^j) - (r_{t}^j + \gamma \max_{a_{t+1}^j} Q(\mathbf{\Theta}_{t+1}^j,a_{t+1}^j)) \right \|^2,
\end{equation}
where $Q(\mathbf{\Theta}_t^j,a_t^j)$ is the Q-value function;  $\gamma$ is the discount factor applied to future rewards.

\subsubsection*{End-to-end fine-tuning using weighted losses}
Aiming for a better hand-eye coordination, an end-to-end fine-tuning is conducted for a combined network (perception + control) after their separate training, using weighted task ($L_q$) and perception ($L_p$) losses. 
Here for end-to-end fine-tuning, $\mathbf{\Theta}^j$ is replaced with $I^j$ in $L_q$ (Eq.~\ref{equ:control_cost}). The control network is updated using only $L_q$, while the perception network is updated using the weighted loss 
\begin{equation}
\label{equ:endtoend_cost}
L= \beta L_p + (1-\beta) L_q^{BN},
\end{equation}
where $L_q^{BN}$ is a pseudo-loss which reflects the loss of $L_q$ in the bottleneck (BN); $\beta \in [0,1]$ is a balancing weight. From the backpropagation algorithm~\cite{lecun-88}, we can infer that $\delta_L = \beta \delta_{L_p} + (1-\beta) \delta_{L_q^{BN}} $, where $\delta_L$ is the gradients resulted by $L$; $\delta_{L_p}$ and $\delta_{L_q^{BN}}$ are the gradients resulting  respectively from $L_p$ and $L_q^{BN}$ (equivalent to that resulting from $L_q$ in the perception module).

\section{Benchmark: Robotic Reaching}
\label{sec:benchmark}

We use the canonical planar reaching task in~\cite{zhang2015towards} as a benchmark to evaluate the feasibility of the modular DQN and its training method. The task is defined as controlling a robot arm so that its end-effector position $\mathbf{x}$ in operational space moves to the position of a target $\mathbf{x}^* \in \mathbb{R}^m$. The robot's joint configuration is represented by its joint angles $\mathbf{q} \in \mathbb{R}^n$. The two spaces are related by the forward kinematics, i.e., $\mathbf{x}=\mathcal{K}(\mathbf{q})$.
The reaching controller adjusts the robot configuration to minimize the error between the robot's current and target position%
, i.e., $\left \| \mathbf{x} - \mathbf{x}^*\right \| $. In this task, we use the target position $\mathbf{x^*}$ and arm configuration $\mathbf{q}$ to represent the scene configuration $\mathbf{\Theta}$. 
The physical meaning of $\mathbf{\Theta}$ guarantees the convenience of collecting labelled training data, as it can directly be measured.
%
We consider a robotic arm (Figure~\ref{fig:sim_to_real}) with 3 degrees of freedom (DoF), i.e., $\mathbf{q} \in \mathbb{R}^3$ steering its end-effector position in the plane i.e., $\mathbf{x} \in \mathbb{R}^2$ -- ignoring orientation.


\subsubsection{Task setup}
\label{sec:problem}
\begin{figure}[tpb!]
    \centering
    \includegraphics[width=1.0\columnwidth]{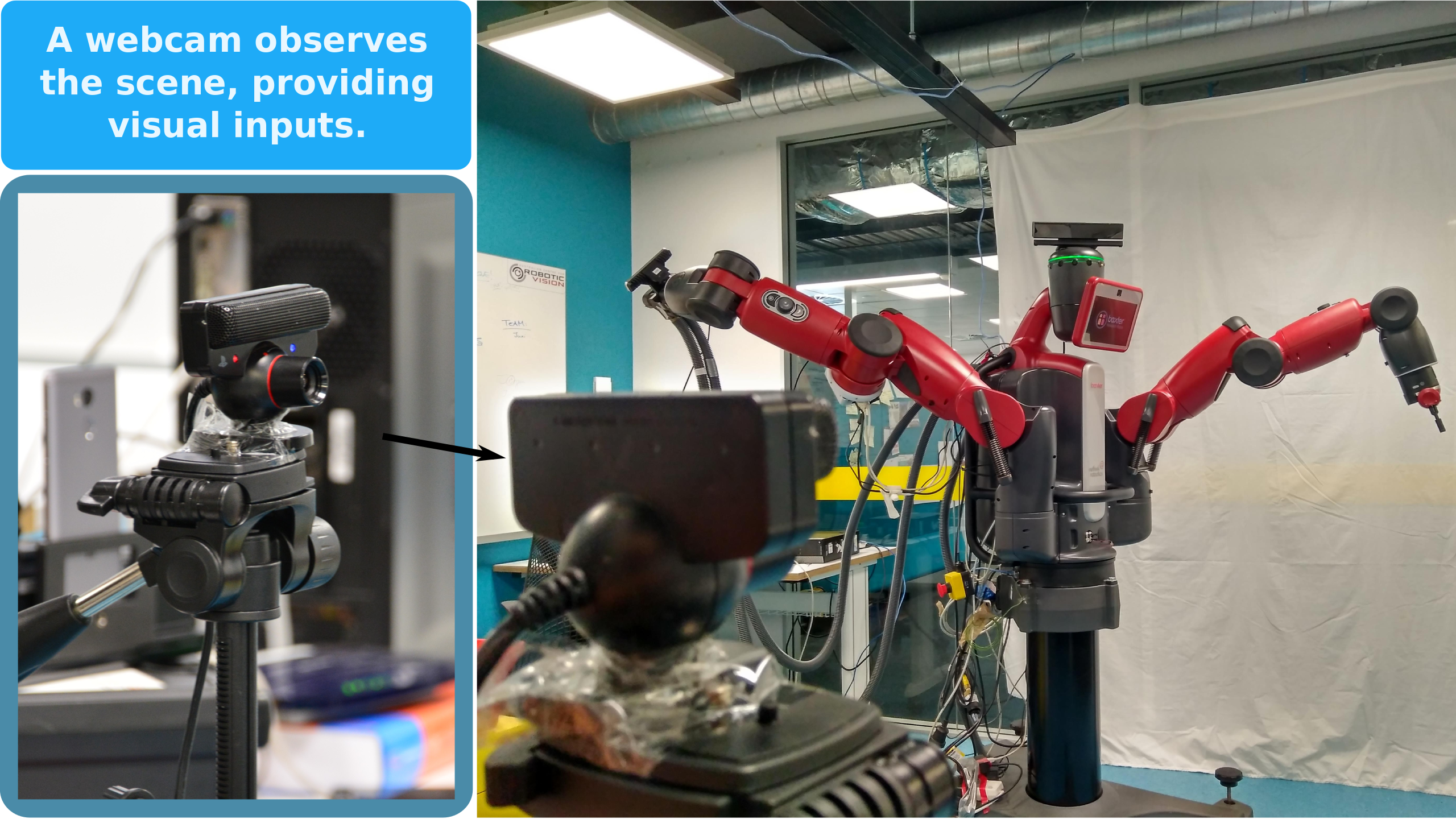}
    \caption{A webcam is used to observe the scene, providing visual inputs.}
    \label{fig:setup}
\end{figure}

The real-world task employs a Baxter robot's left arm  (Figure~\ref{fig:sim_to_real}B) to reach (in a vertical plane) for an arbitrarily placed blue target using vision.
We control only three joints, keeping the others fixed.  At each time step one of 9 possible actions $a \in \mathbf{a}$ is chosen to change the robot configuration, 3 per joint: increasing or decreasing by a constant amount (0.04\unit{rad}) or leaving it unchanged. 
A monocular webcam is placed on a tripod to observe the scene, providing raw-pixel image inputs (Figure~\ref{fig:setup}).

\subsubsection{Simulator}
A simple simulator was created that, given a scene configuration $\mathbf{\Theta} = [\mathbf{x^*} \, \mathbf{q}] \in \mathbb{R}^5$, generates the corresponding image.
It creates images using a simplistic representation of a Baxter arm (in configuration $\mathbf{q}$) and the target (at location $\mathbf{x^*}$) represented by a blue disc with a radius of 3.8\unit{pixels} (9\unit{cm}) in the image (Figure~\ref{fig:sim_to_real}A).
A reach is deemed successful if the robot reaches and keeps its end-effector within 7\unit{pixels} (16\unit{cm}) of the target's centre for four consecutive actions. Experimental results show that although the simulator is low-fidelity, and therefore cheap and fast for data collection, reaching skills can be learned and transferred to the real robot.

\subsubsection{Network architecture}
\label{sec:meth_perception}
The perception network for the task has an architecture as shown in Figure~\ref{fig:sim_to_real}C, which consists of 3 convolutional and 1 fully-connected (FC) layer. Images from the simulator or the webcam (RGB, cropped to $160\times210$) are converted to grey-scale and downsized to $84\times84$ as inputs to the network. 

The control network consists of 3 fully-connected layers, with 400 and 300 units in the two hidden layers (Figure~\ref{fig:sim_to_real}C). Input to the control network is the scene configuration $\mathbf{\Theta}$, its outputs are the Q-value estimates for each of the 9 possible actions.
%

Networks with a first convolutional layer initialized with weights from pre-trained GoogLeNet~\cite{szegedy2015going} (on ImageNet data~\cite{deng2009imagenet}) were observed to converge faster and achieve higher accuracy.
As GoogLeNet has three input channels (RGB) compared to our single (grey) channel network a weight conversion, based on standard RGB to grey-scale mapping, is necessary in the first convolutional layer initialization.
The other parts of the networks are initialized with random weights.

\subsubsection{Reward}
The reward $r$ for Q learning is determined by the Euclidean distance $d=\left \| \mathbf{x} - \mathbf{x}^*\right \| $ between the end-effector and the target disc's centre 
\begin{equation}
\label{equ:reward_function}
r= \begin{cases}
 \lambda(\delta/d-1), & \text{if}\ d>\delta\\ 
 0, &  \text{if}\ d\leqslant \delta,\ n<N\\ 
 1, &  \text{if}\ d\leqslant \delta,\ n\geqslant N
\end{cases}
\end{equation}
where $\delta$ is a threshold for reaching a target ($\delta=0.05$m); $\lambda$ is a constant discount factor ($\lambda=10^{-3}$); $n$ represents the times of $d$ is consecutively smaller than $\delta$ and $N=4$ is a threshold that determines task completion. 
This reward function will yield negative rewards until getting close enough to the target. 
This helps to take into account temporal costs during policy search, i.e., fewer steps are better.
By giving positive rewards only when $d$ is smaller than the threshold $\delta$ for more than $N$ consecutive times, the reward function will guide a learner to converge to a target rather than just pass through it. This reward function proves successful for learning planar reaching, 
but we do not claim optimality. Designing a good reward function is an active topic in reinforcement learning.

\subsubsection{Guiding Q learning with K-GPS}
In Q-learning, the $\varepsilon$-Greedy method is frequently used for policy search. 
However, our experiments show that $\varepsilon$-Greedy works poorly for the planar reaching task when using multiple DoF (Section~\ref{sec:control}). Therefore, we introduce a kinematics-based controller to guide the policy search (K-GPS), i.e., guide the learning of the operational space controller with a joint-space controller, which 
selects actions by
\begin{equation}
\label{equ:k_gps}
\arg \underset{a}{\min}\ \left \| a[\mathbf{q}] - \mathcal{K}^{-1}(\mathbf{x}^*) \right \|,
\end{equation}
where $a[]$ is an operator that returns an updated arm configuration when executing an action, and $\mathcal{K}^{-1}(\cdot)$ is the inverse kinematic function.


\begin{algorithm}[h]
Initialize replay memory $\mathbf{D}$\\
Initialize Q-function $Q(\mathbf{\Theta},a)$ with random weights\\
\For{iteration=1,K}{
	\If{previous trial finished}{
	Start a new trial:\\
	Randomly generate configurations $\mathbf{q}$ and $\mathbf{q}^*$\\
	Compute the end-effector position $\mathbf{x}^* =\mathcal{K}(\mathbf{q}^*)$\\
	}
	\textbf{if} \emph{rand(0,1)} $<\varepsilon$ \textbf{then}
		$a_t =\arg \underset{a}{\min}\ \left \| a[\mathbf{q}] - \mathbf{q}^* \right \|$\\
	~ ~~ ~ ~ ~ ~ ~ ~ ~ ~\textbf{else}  $a_t = \arg \underset{a}{\max}\ Q(\mathbf{\Theta}_t,a)$ \\
	Execute $a_t$ and observe $r_t$ and $\mathbf{\Theta}_{t+1}$\\
	Add the new sample ($\mathbf{\Theta}_t,a_t,r_t,\mathbf{\Theta}_{t+1}$) into $\mathbf{D}$\\
	Sample a random mini-batch from $\mathbf{D}$\\
	Update ($Q(\mathbf{\Theta},a)$) using the mini-batch  
}
\caption{DQN with K-GPS}
\label{alg:k_gps}
\end{algorithm}%

Algorithm~\ref{alg:k_gps} shows the DQN with K-GPS.
A replay memory $\mathbf{D}$ is used to store samples of ($\mathbf{\Theta}_t,a_t,r_t,\mathbf{\Theta}_{t+1}$). At the beginning of each trial, the arm's starting configuration and target position are randomly generated. To guarantee a random target position $\mathbf{x}^*$ is reachable by the arm, we first randomly select an arm configuration $\mathbf{q}^*$, then use the position of its end-effector as the target position. $\mathbf{q}^*$ ($=\mathcal{K}^{-1}(\mathbf{x}^*)$) is also used as the desired configuration to guide the policy search. In each iteration, the action will be selected either by the kinematic controller (with probability $\varepsilon$) or by the control network. During training, $\varepsilon$ decreases linearly from 1 to 0.1, i.e., the guidance gets weaker in the process. 
The newly observed sample ($\mathbf{\Theta}_t,a_t,r_t,\mathbf{\Theta}_{t+1}$) is added to $\mathbf{D}$ before the network is updated using a mini-batch randomly selected from $\mathbf{D}$.
\section{Experiments and Results}
\label{sec:experiments}


Perception and control networks were first trained and evaluated independently under various conditions for the benchmark reaching task. Then comparisons were made for different combined (end-to-end) networks, such as naively combined vs fine-tuned networks. Evaluations were conducted in both simulated and real scenarios: a Baxter robot arm reaching observed by a camera.

\subsection{Assessing Robot Perception}
\label{sec:perception_experiments}
To understand the effect of adapting perception with real images, we trained six networks for the planar reaching task with different training data as shown in Table~\ref{tab:perception_experiments_conditions}. \textbf{SIM} was trained from scratch purely in simulation; \textbf{RW} was trained from scratch using real images; \textbf{P25}--\textbf{100} were trained by adapting \textbf{SIM} with different percentages of real images found in the mini-batches.

\begin{table*}[tb]
\renewcommand\arraystretch{1.2}
\renewcommand\tabcolsep{6pt}
 \caption{Perception Networks, Conditions and Error Reported}
 \label{tab:perception_experiments_conditions}
 \label{tab:perception_experiments_results}
 \centering	
 \begin{tabular}{c | l | c | c | c | c | c | c}	

 \toprule
 
 \multirow{2}{*}{\bfseries Nets} & 
 \multirow{2}{*}{\bfseries Training Conditions} &  
 \multicolumn{2}{c|}{\bfseries Sim} &
 \multicolumn{2}{c|}{\bfseries Real} &
 \multicolumn{2}{c}{\bfseries Live} 
 \\ 
 
  & & $e_{\mu}$ & $e_{\sigma}$ & $e_{\mu}$ & $e_{\sigma}$ & $e_{\mu}$ & $e_{\sigma}$\\ 

 \hline 
 \textbf{SIM} & Train from scratch, simulated images    & 0.013 & 0.009 & 13.92 & 0.877 & 13.53 & 1.436\\
 \textbf{RW}  & Train from scratch, real images    & 0.537 & 0.191 & 0.023 & 0.046 & 0.308 & 0.138\\
 \hline
 \textbf{P25}  & Adapt \textbf{SIM}, 25\% real, 75\% simulated images & \bfseries 0.012 & 0.008 & 0.025 & 0.044 & 0.219 & 0.091\\
 \textbf{P50}  & Adapt \textbf{SIM}, 50\% real, 50\% simulated images & 0.013 & 0.008 & 0.024 & 0.045 & 0.192 & 0.109\\
 \textbf{P75}  & Adapt \textbf{SIM}, 75\% real, 25\% simulated images & 0.015 & 0.010 & 0.021 & 0.046 & 0.135 & 0.123\\
 \textbf{P100} & Adapt \textbf{SIM}, 100\% real images    & 0.498 & 0.162 & \bfseries 0.019 & 0.049 & \bfseries 0.133 & 0.153\\


 \bottomrule 
 \end{tabular}
 \end{table*}

\begin{figure}[tpb!]
    \centering
    \includegraphics[width=1.0\columnwidth]{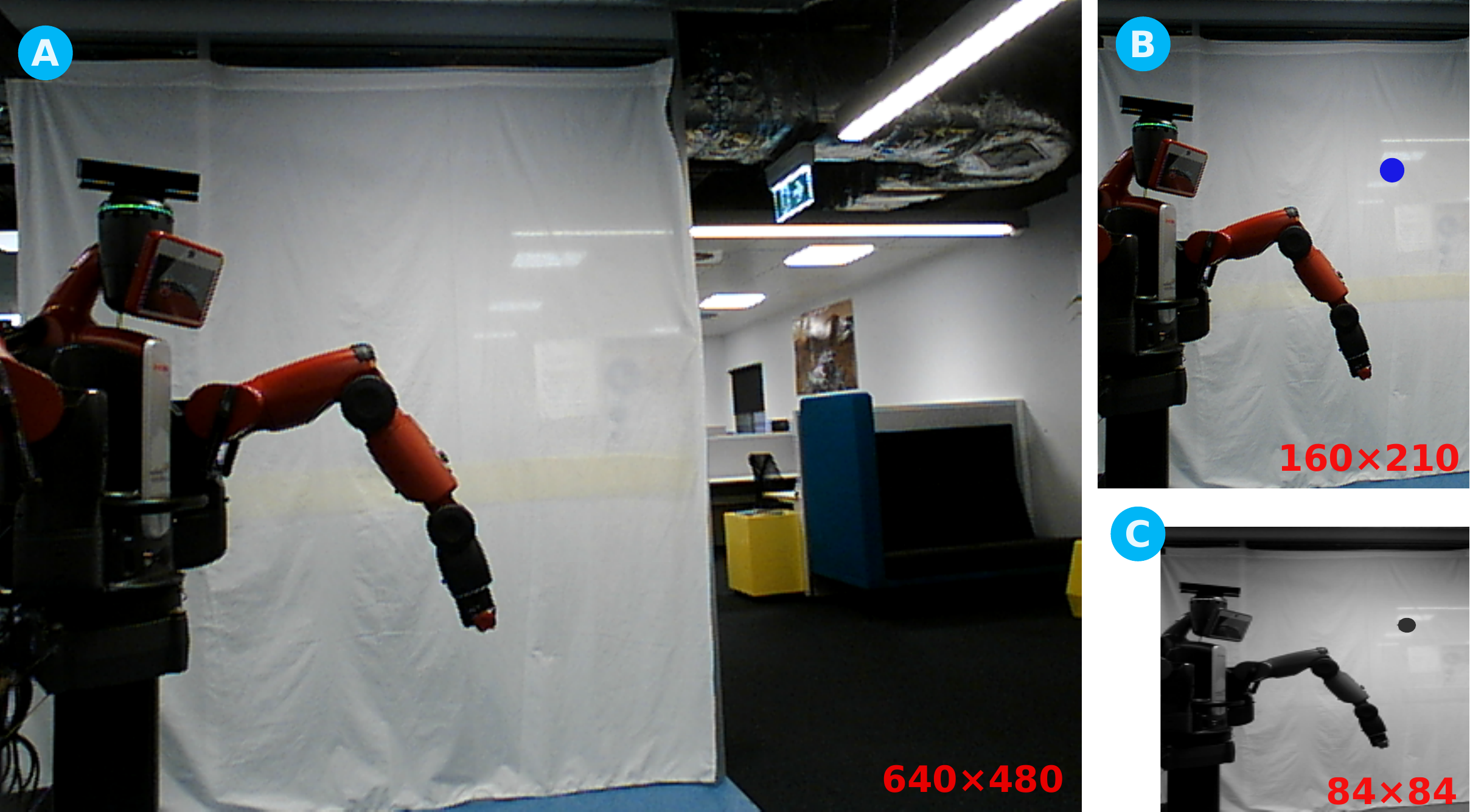}
    \caption{An image from a webcam (A) is first cropped and scaled to match the simulator size, a virtual target is also added (B). Like the simulated images, it is then converted to grey-scale and scaled to $84\times84$ (C).}
    \label{fig:real_sample}
\end{figure}

1418 images were collected on the real robot together with their ground-truth scene configuration for use in training and adaptation.
During image collection, the robot was moved to fixed arm configurations uniformly distributed in the joint space.
The target (blue disc) is rendered into the image at a random position to create a large number of training samples.
Figure~\ref{fig:real_sample} shows a typical scene during data capture and a final dataset image after scaling, cropping and target addition.
To increase the robustness of the trained network the image dataset was augmented by applying transformations to the original images (rotation, translation, noise, and brightness). 


In both training and adaptation, RmsProp~\cite{tieleman2012lecture} was adopted using a mini-batch size of 128 and a learning rate between 0.1 and 0.01. The networks trained from scratch converged after 4 million update steps (\textasciitilde 6 days on a Tesla K40 GPU). In contrast, those adapted from \textbf{SIM} converged after only 2 million update steps. 

Performance was evaluated using perception error, defined as the Euclidean norm between the predicted and ground-truth scene configuration $e= \left \| \widehat{\mathbf{\Theta}} - \mathbf{\Theta}^* \right \|$. We compared three different scenarios: \\
\textbf{[Sim]} 400 simulator images, uniformly distributed in the scene configuration space\\
\textbf{[Real]} 400 images collected using the real robot but withheld during training\\ 
\textbf{[Live]} 40 different scene configurations during live trials on Baxter

\begin{figure}[tpb!]
    \centering
    \includegraphics[width=1.0\columnwidth]{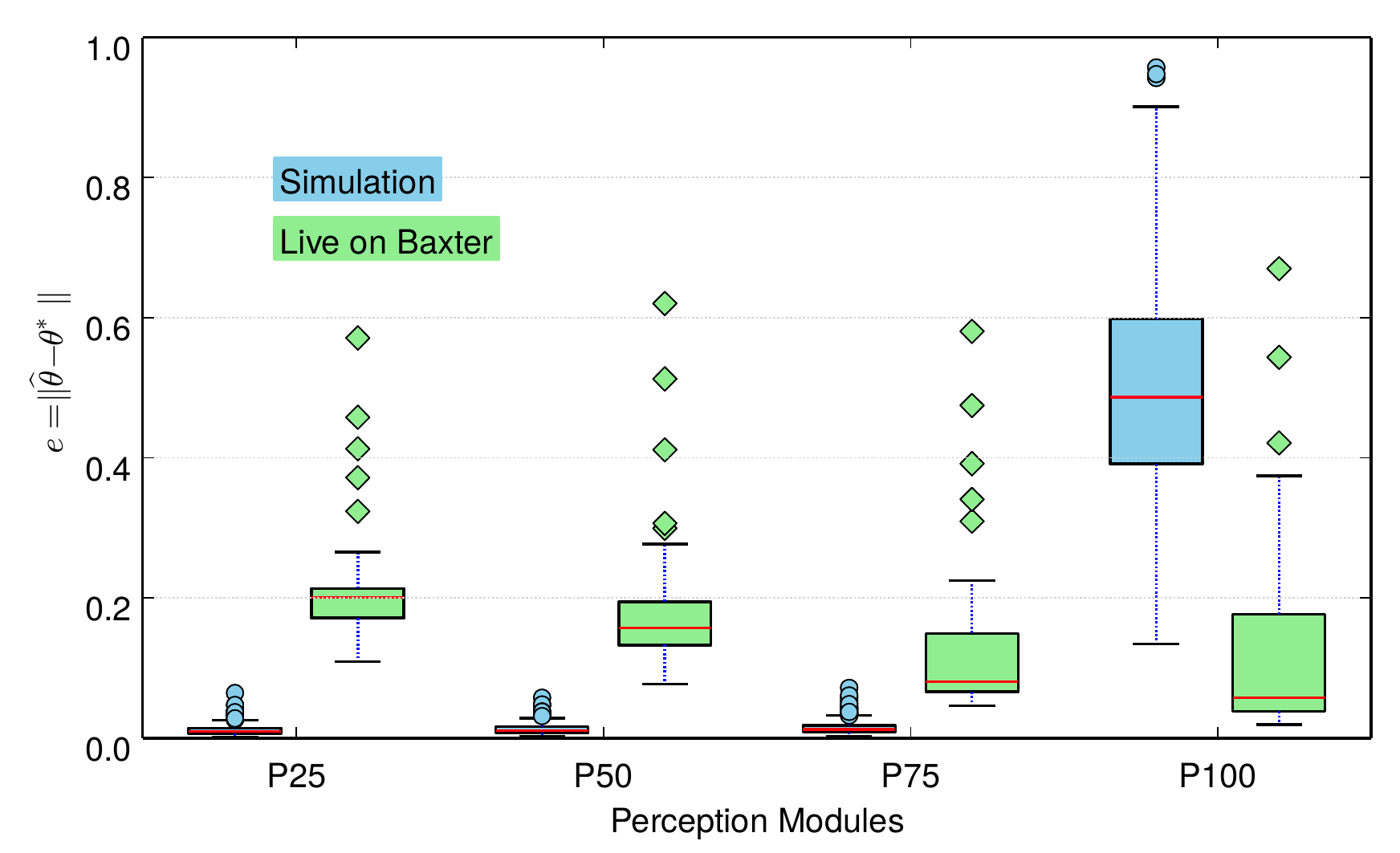}
    \caption{Distance errors for networks \textbf{P25} to \textbf{P100} in simulation (blue) and during live trials on Baxter (green). The  circles and diamonds represent outliers.}
    \label{fig:perception_results}
\end{figure}

Results are listed in Table~\ref{tab:perception_experiments_results} with mean $e_{\mu}$ and standard deviation $e_{\sigma}$ of $e$.
As expected, the perception network performed well in the scenarios in which they were trained or adapted but poorly otherwise. 
The network trained with only simulated images (\textbf{SIM}) had a small error in simulation but very poor performance in real scenarios (Real and Live). Similarly, the network trained (\textbf{RW}) or adapted (\textbf{P100}) with only real images performed fine in real scenarios but poorly in simulation.
In contrast, the networks adapted with a mixture of simulated and real images coped with all scenarios.

Results for \textbf{P25} to \textbf{P100} show that the fraction of real images in a mini-batch is important for balancing real and simulated environment performance (Figure~\ref{fig:perception_results}).
The more real images presented 
during training the smaller the real world experiment error became -- similarly for simulation.
In particular, \textbf{P25} had the smallest mean error $e_{\mu}$ in simulation and \textbf{P100} the smallest $e_{\mu}$ for real world and live scenarios. However, when balancing $e_{\mu}$ and $e_{\sigma}$, \textbf{P75} had the best performance when tested live on Baxter: it had a smaller $e_{\sigma}$ and only slightly larger $e_{\mu}$ compared to \textbf{P100}.

Comparing the performance 
in simulation we see that the network adapted with no simulated images  (\textbf{P100} in Figure~\ref{fig:perception_results}) resulted in a much larger error than \textbf{SIM}. This indicates that the presence of simulated images in adaptation prevents a network from forgetting the skills learned. 
We  also observe that, a network adapted using only real images (\textbf{P100}) had a smaller error than one trained from scratch (\textbf{RW}). This shows that adaptation from a pre-trained network leads to better performance as well as reduces the training time.


For all networks except \textbf{SIM}, errors in live trials on Baxter were slightly larger than that for the real world testing set, although the collected real world dataset was augmented with translations and rotations in training. This indicates a high sensitivity of the perception networks to variations in camera pose (between capture of the training/testing images and the live trials).  
To further test this indication we trained some perception networks \textit{without} data augmentation, which resulted in significantly poorer performance during live trials.


To check sensible network behaviour, we investigated the perception networks behaviour when no target was present.
All trained networks output incorrect constant values (with small variance) for the target position prediction.
When images with two targets were presented to the networks, a random mixture of the two target positions were output.
However in both cases, joint angles were estimated accurately. When part of the robot body or arm was occluded, as shown in Figure~\ref{fig:real_collision}, the arm configurations were still estimated well, although with a slightly greater error in most cases. 


\subsection{Assessing Robot Control}
\label{sec:control}

\begin{table*}[tb!]
\caption{Performance of $\varepsilon$-Greedy and K-GPS}
\label{tab:gps_experiments_results}
\centering	
\renewcommand\arraystretch{1.2}
\renewcommand\tabcolsep{6pt}
\begin{tabular}{c | c | c | c | c | c | c }	
\toprule 

\multirow{2}{*}{DoF} & 

\multicolumn{2}{c|}{$d_{\mbox{med}}$ [cm]} & \multicolumn{2}{c|}{$d_{Q3}$ [cm]} &
\multicolumn{2}{c}{$\alpha$ [\%]}
\\ 

\ & $\varepsilon$-Greedy & K-GPS & $\varepsilon$-Greedy & K-GPS & $\varepsilon$-Greedy & K-GPS\\ 

\hline 
1 & 1.0 & \bfseries 0.7 & 2.3 & \bfseries 1.1 & \bfseries 100.00 & \bfseries 100.00 \\
2 & 4.9  & \bfseries 2.8 & 9.0 & \bfseries 3.7 & 83.75 & \bfseries 99.50 \\
3 & 14.5 & \bfseries 3.4 & 28.5 & \bfseries 4.3 & 50.25 & \bfseries 98.50 \\

\bottomrule 
\end{tabular}
\end{table*}

We trained 6 control networks in simulation for the planar reaching task with varying degrees of freedom using $\varepsilon$-Greedy or K-GPS policy search. 
In the 1 DoF case, only $q_2$ was active; 2 DoF uses $q_2$ and $q_3$; while 3 DoF controls all three joints.
In training, we used a learning rate between 0.1 and 0.01, and a mini-batch size of 64. The $\varepsilon$ probability decreased from 1 to 0.1 within 1 million training steps for 1 DoF reaching, 2 million steps for 2 DoF, and 3 million steps for 3 DoF. $\varepsilon$-Greedy and K-GPS used the same $\varepsilon$. 

\begin{figure}[tpb!]
    \centering
    \includegraphics[width=1.0\columnwidth]{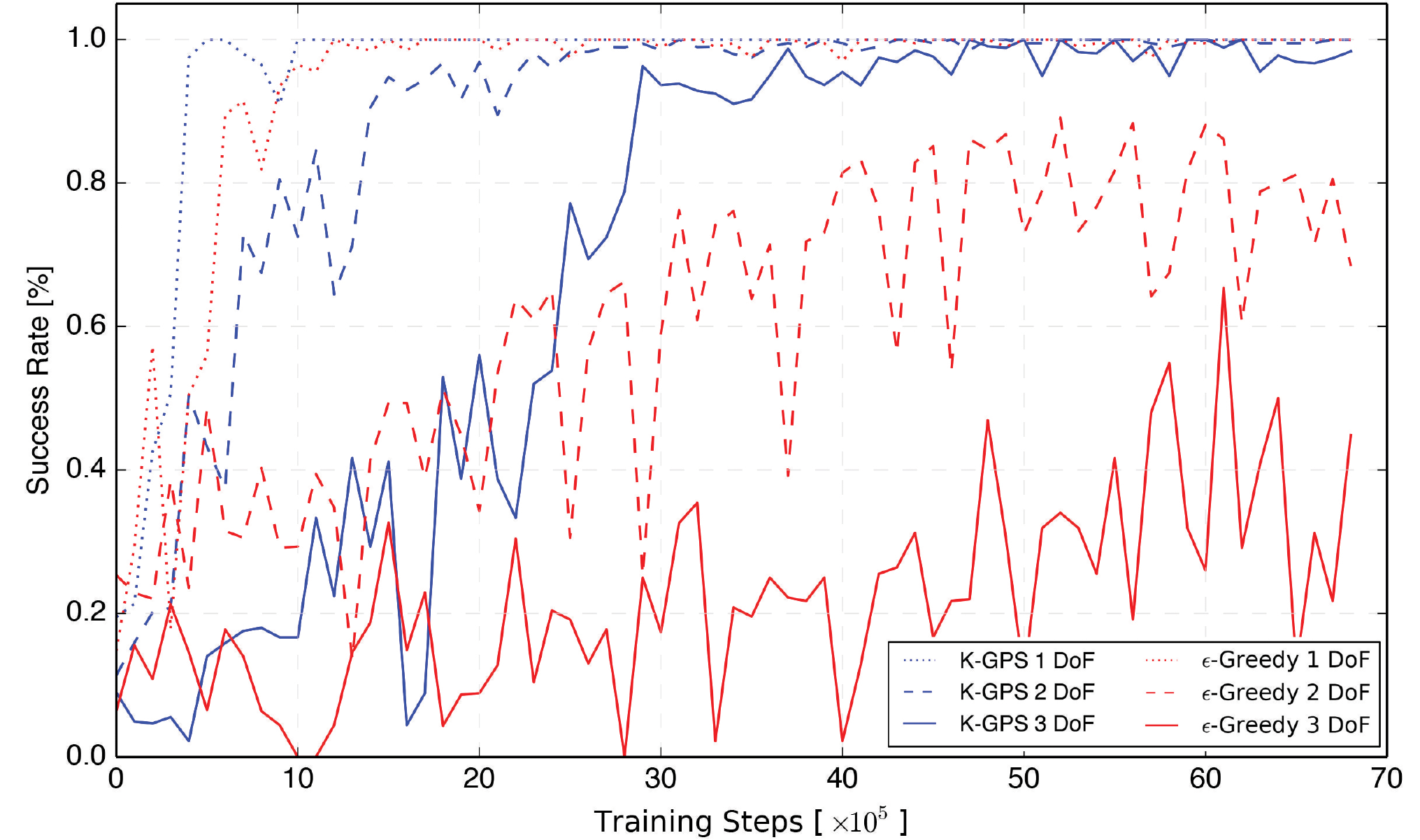}
    \caption{Learning curves showing that K-GPS converges faster than $\varepsilon$-Greedy.}
    \label{fig:control_learning_curve}
\end{figure}

Figure~\ref{fig:control_learning_curve} shows the learning curves indicating the success rate of a network after a certain number of training steps.
For each data point 200 reaching tests were performed using the criteria introduced in Section~\ref{sec:problem}.
The target positions and initial arm configurations were uniformly distributed in the scene configuration space.

For 1 DoF reaching, networks trained using K-GPS and $\varepsilon$-Greedy both converged to a success rate of 100\% after around 1 million steps (4 hrs on one 2.66GHz 64bit Intel Xeon processor core). For the 2 DoF case, K-GPS and $\varepsilon$-Greedy converged to around 100\% and 80\% and took 2 million (8 hrs) and 4 million (16 hrs) steps respectively. For 3 DoF reaching, they converged to about 100\% and 40\% after 4 million and 6 million (24 hrs) steps respectively. The results show that K-GPS was feasible for all degrees of freedom, while $\varepsilon$-Greedy worked appropriately only in 1 DoF reaching and degraded as the number of DoF increased.

For a more detailed comparison, we further analyzed the error distance $d$ -- the Euclidean distance between the end-effector and target -- reached by a converged network. 
400 reaching tests were performed for each network in simulation.
The results are sumarized in Table~\ref{tab:gps_experiments_results} which shows that K-GPS achieved smaller error distances for both median $d_{\mbox{med}}$ and third quartile $d_{Q3}$ than $\varepsilon$-Greedy in all DoF cases.

To evaluate the performance of a control network in real scenarios, a K-GPS trained network (3 DoF) was directly transferred on Baxter.
In the test, joint angles were taken from the robot's encoders and the target position was set externally.
It achieved a median distance error of 1.3\unit{pixels} (3.2\unit{cm}) in 20 consecutive reaching trials (CR in Table~\ref{tab:e2e_experiments_results}), indicating robustness to real-world sensing noise.




In addition to the proposed FC network architecture, we also tested several other control network architectures, varying the number of hidden layers and the number of units in each layer. Qualitative results show that a network with only one hidden layer was enough for 1 DoF reaching but insufficient for 2 and 3 DoF cases. The number of units in each layer also influenced the performance.
Our proposed architecture worked best for 3 DoF reaching;
at least two hidden layers with 200 and 150 units were needed.


\subsection{End-to-end Network Performance}
\label{sec:e2e_reaching}


We evaluated the end-to-end performance of combined networks in both simulated and real-world scenarios using the metrics of Euclidean distance error $d$ (between the end-effector and target) and average accumulated reward $\bar{R}$ (a bigger accumulated reward means a faster and closer reaching to a target) in 400 simulated trials or 20 real trials. When testing in real scenarios, virtual targets were rendered into the image stream from the camera for repeatability and simplicity.

For comparison, we evaluated three networks end-to-end: EE1, EE2 and EE2-FT. EE1 is a combined network comprising \textbf{SIM} and CR; EE2 consists of \textbf{P75} and CR; EE2-FT is EE2 after end-to-end fine-tuning using weighted losses. 
\textbf{P75} and CR are the perception and control modules selected in Section~\ref{sec:perception_experiments} and Section~\ref{sec:control}, which have the best performance individually.

The end-to-end fine-tuning was mainly conducted in simulation. In the fine-tuning, $\beta = 0.8$, we used a learning rate between 0.01 and 0.001, a mini-batch size of 64 and 256 for task and perception losses respectively, and an exploration possibility of 0.1 for K-GPS. These parameters were empirically selected.  
To prevent the perception module from forgetting the skills for real scenarios, the 1418 real samples were also used to obtain $\delta_{L_p}$. Similar to \textbf{P75}, 75\% samples in a mini-batch for $\delta_{L_p}$ were from real scenarios, i.e., at each weight updating step, 192 real and 64 simulated samples were used.

\begin{table*}[tb!]
\caption{End-to-end Reaching Performance}
\label{tab:e2e_experiments_results}
\centering	
\renewcommand\arraystretch{1.2}
\renewcommand\tabcolsep{6pt}
\begin{tabular}{c | c | c | c | c | c | c }	
\toprule 

\multirow{2}{*}{Scenario} & \multirow{2}{*}{Nets} & 
\multicolumn{2}{c|}{$d_{\mbox{med}}$} &
\multicolumn{2}{c|}{$d_{Q3}$} & $\bar{R}$\\

\ & \ &  [cm] &  [pixels] & [cm] &  [pixels] &  [\textbackslash] \\ 

\hline 
\multirow{4}{*}{Sim} & EE1 (\textbf{SIM}+CR) & 4.7 & 2.0 & 6.7 & 2.8 & 0.313\\
\ & EE2 (\textbf{P75}+CR)  & 4.6 & 1.9 & 6.2 & 2.6 & 0.319 \\
\ & \bfseries EE2-FT  & \bfseries 3.6 & \bfseries 1.5 & \bfseries 4.8 & \bfseries 2.0 & \bfseries 0.626 \\
\ & \emph{CR (Control Only Baseline)}  & \emph{3.4} & \emph{1.4} & \emph{4.3} & \emph{1.8} &  \emph{0.761} \\
\hline
\multirow{4}{*}{Real} & EE1 & 41.8 & 17.5 & 80.2 & 33.6 & -0.050\\
\ & EE2 & 4.6 & 1.9 & 6.2 & 2.6 & 0.219\\
\ & \bfseries EE2-FT & \bfseries 3.7 & \bfseries 1.6 & \bfseries 5.2 & \bfseries 2.2 & \bfseries 0.628\\
\ & \emph{CR (Control Only Baseline)}  & \emph{3.2} &  \emph{1.3} & \emph{4.3} & \emph{1.8} & \emph{0.781}\\


\bottomrule 
\end{tabular}
\end{table*}

\begin{figure*}[tpb!]
\begin{center}
\includegraphics[width=2.0\columnwidth]{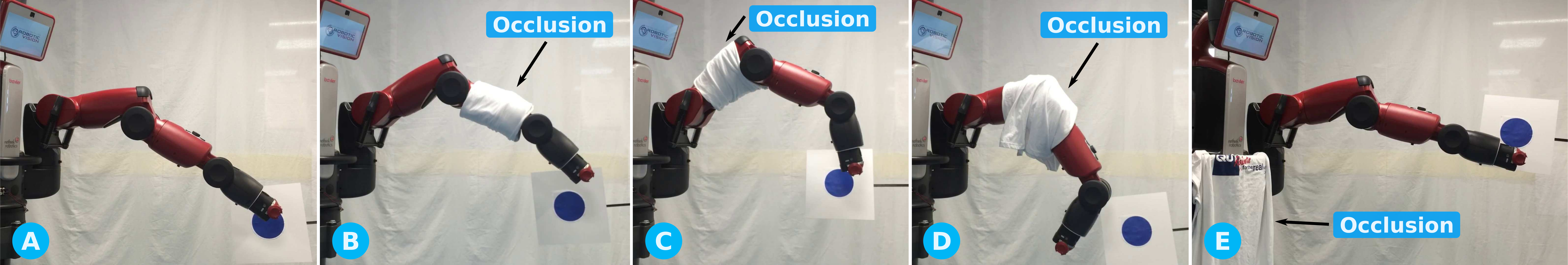}
\end{center}
\caption{Successful reaching with real targets (A) and occlusions (B-E) which were not present during training. }
\label{fig:real_collision}
\end{figure*}

The error distances in \unit{cm} and \unit{pixels} (in the $84\times84$ image) are compared. Results are listed in Table~\ref{tab:e2e_experiments_results}, where $d_{\mbox{med}}$ and $d_{Q3}$ are the median and third quartile of $d$. The CR network, where perfect perception is assumed is added as baseline.

From the results in simulation, we can see that EE1 and EE2 have similar performance in all metrics. After end-to-end fine-tuning, EE2-FT achieved a much better performance (21.7\% smaller $d_{\mbox{med}}$ and 96.2\% bigger $\bar{R}$) than EE2.
The fine-tuned performance is very close to that of the control module (CR) which controls the arm using ground-truth $\mathbf{\Theta}$ as sensing inputs. This indicates the proposed fine-tuning approach significantly improved the hand-eye coordination.

In the real world, as expected, EE1 worked poorly, since the perception network had not experienced real scenarios. In contrast, EE2 and EE2-FT worked well and achieved comparable performance to that in simulation. Note that due to the cost of real world experiments, only 20 trials each were run (compared to 400 in simulation). Similar to the results in simulation, benefiting from the end-to-end fine-tuning, EE2-FT achieved a smaller median distance error (3.7\unit{cm}, 1.6\unit{pixels}) than EE2. This shows that the adaptation to real scenarios can be kept by presenting (a mix of simulated and) real samples to compute the perception loss. All networks (except EE1 in the real scenario) achieved a success rate between 98\% and 100\%.

Apart from the weighted end-to-end fine-tuning approach, we also tried naively fine-tuning combined networks only using the task loss $L_q$. It did not work well for performance improvement (making the performance even worse), although many efforts were made in searching appropriate hyper-parameters.

To see the combined networks' robustness to a real target and occlusions, we tested EE2-FT in the setups shown in Figure~\ref{fig:real_collision}. In the case without occlusion (A), real targets can be reached (although only virtual targets were used for training).
Occlusions had not been experienced by the network during training, yet we see that in cases B, C and E, most targets can be reached but with larger distance 
errors (about 2 times larger than in case A).
In case D, only a few targets could be reached with a yet increased error across all cases, as shown in the attached video\footnote{\url{https://goo.gl/vtLuVV}}.

\section{Conclusion}

In this paper, we demonstrated reliable vision-based planar reaching on a real robot using a modular deep Q network (DQN), trained in simulation, with transference to a real robot utilizing only a small number of real world images.
The proposed end-to-end fine-tuning approach using weighted losses significantly improved the hand-eye coordination of the naively combined network (EE2). Through fine-tuning, its (EE2-FT) reaching accuracy was improved by 21.7\%. This work has led to the following observations:




\subsubsection{Value of a modular structure and end-to-end fine-tuning:}
The significant performance improvement (hand-eye coordination) and relatively low real world data requirements show the feasibility of the modular structure and end-to-end fine-tuning for low-cost transfer of visuo-motor policies. Through fine-tuning with weighted losses, a combined network comprising perception and control modules trained independently can even achieve performance very close to the control network alone (indicating the performance upper-limit). Scaling the proposed techniques to more complicated tasks and networks will likely be achievable with an appropriate scene configuration representation.


\subsubsection{Perception adaptation:}
A small number of real-world images are sufficient to adapt a pre-trained perception network from simulated to real scenarios in the benchmark task, even with a simulator of only modest visual fidelity. The percentage of real images in a mini-batch plays a role in balancing the performance in real and simulated environments. The presence of simulated images in fine-tuning prevents a network from forgetting pre-mastered skills. The adapted perception network also has some interesting robustness properties: it can still estimate the robot configuration even in the presence of occlusions it has not directly experienced or when there is/are zero or multiple targets.

\subsubsection{Control training with K-GPS:}
With  guidance from a kinematic controller K-GPS leads to better policies (smaller error distance) in a shorter time than $\varepsilon$-Greedy, producing a trained control network that is robust to real-world sensing noise.
However K-GPS does assume that we already have some knowledge of the task to learn, i.e., a model of the task.

We believe the architecture presented here: introducing a bottleneck between perception and control, training networks independently then merging and fine-tuning, is a promising line of investigation for robotic visual servoing and manipulation tasks. In current and future work we are scaling up the complexity of the robot tasks and further characterizing the performance of this approach. 
Promising results have been obtained in table-top object reaching in clutter using a 7 DoF robotic arm in velocity control mode~\cite{zhang2018icra}.

\section*{Acknowledgements}

This research was conducted by the Australian Research Council Centre of Excellence for Robotic Vision (project number CE140100016). Additional computational resources and services were provided by the HPC and Research Support Group at the Queensland University of Technology, Brisbane, Australia.

\bibliographystyle{named}
\bibliography{rl,nc,fangyi}  

\end{document}